\documentclass[journal]{IEEEtran}
\usepackage{amsmath,amsfonts}
\usepackage{algorithmic}
\usepackage{array}
\usepackage[caption=false,font=normalsize,labelfont=sf,textfont=sf]{subfig}
\usepackage{textcomp}
\usepackage{stfloats}
\usepackage{url}
\usepackage{verbatim}
\usepackage{graphicx}
\usepackage{xcolor}
\usepackage{tabularx}
\def\BibTeX{{\rm B\kern-.05em{\sc i\kern-.025em b}\kern-.08em
    T\kern-.1667em\lower.7ex\hbox{E}\kern-.125emX}}
\usepackage{balance}

\usepackage[normalem]{ulem}

\newcommand{\sectionref}[1]{Section\,\ref{#1}}
\newcommand{\figureref}[1]{Figure\,\ref{#1}}
\newcommand{\tableref}[1]{Table\,\ref{#1}}
\newcommand{\equationref}[1]{Eq.\,\ref{#1}}

\title{Vibration of Soft, Twisted Beams for Under-Actuated Quadrupedal Locomotion*
}

\author{Yuhao Jiang$^{1}$, Fuchen Chen$^{2}$, Jamie Paik$^{1}$, and Daniel M. Aukes$^{2}$
\thanks{*This work is supported by the National Science Foundation Grant No. 1935324. \textit{(Corresponding author: Daniel Aukes)}}
\thanks{$^{1}$Reconfigurable Robotics Laboratory, School of Engineering, EPFL, Lausanne, 1015, Switzerland. \{yuhao.jiang, jamie.paik\}@epfl.ch}
\thanks{$^{2}$The School of Manufacturing Systems and Networks, Fulton Schools of Engineering, Arizona State University, Mesa, AZ, 85212, USA. \{fchen65, danaukes\}@asu.edu}      
}

\begin{document}
\maketitle

\begin{abstract}

Under-actuated compliant robotic systems offer a promising approach to mitigating actuation and control challenges by harnessing pre-designed, embodied dynamic behaviors. This paper presents Flix-Walker, a novel, untethered, centimeter-scale quadrupedal robot inspired by compliant under-actuated mechanisms. Flix-Walker employs flexible, helix-shaped beams as legs, which are actuated by vibrations from just two motors to achieve three distinct mobility modes. We analyze the actuation parameters required to generate various locomotion modes through both simulation and prototype experiments. The effects of system and environmental variations on locomotion performance are examined, and we propose a generic metric for selecting control parameters that produce robust and functional motions. Experiments validate the effectiveness and robustness of these actuation parameters within a closed-loop control framework, demonstrating reliable trajectory-tracking and self-navigation capabilities.
\end{abstract}


\maketitle

\section{Introduction}
\label{sec:intro}
Under-actuated, compliant systems exploit structural dynamics to produce complex robotic motions for locomotion and manipulation, while reducing actuation demands. Leveraging these dynamic behaviors diminishes the need for active actuation, lowers controller complexity, reduces actuator count, and simplifies fabrication~\cite{alberto2025, 9244584}.

Legged robots offer superior maneuverability in cluttered terrain compared to wheeled or tracked platforms~\cite{BISWAL20212017, lee_learning_2020}. However, their reliance on intricate actuation and control typically results in larger size, higher power consumption, and increased fabrication costs, limiting their deployment in swarm robotics and confined spaces.

\begin{figure}[t]
    \centering
    \includegraphics[width = \columnwidth]{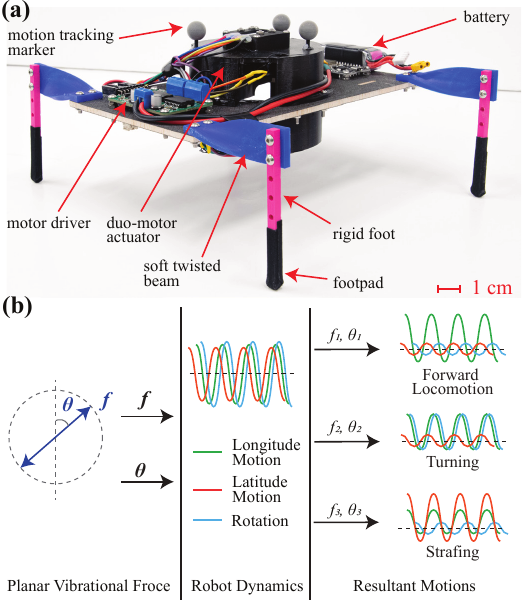}
    \caption{\textbf{Flix-Walker.} (a) The Flix-Walker prototype; (b) the working principle of Flix-Walker: The vibration frequency ($f$) and direction ($\theta$) act as selectors, while the robot’s intrinsic dynamics function as a demultiplexer, selectively expressing specific motion modes. By tuning $f$ and $\theta$, Flix-Walker can achieve three types of mobility: forward locomotion, turning, and strafing.
    }
    \label{fig:concept}
\end{figure}

Modeling and understanding under-actuated systems — specially in the presence of impacts, friction, and environmental uncertainties — remains a significant challenge. In this study, we present Flix-Walker, an untethered, 350\,g quadruped robot that achieves directional locomotion via a controllable vibrational source (\figureref{fig:concept}(a)). Control parameters are optimized for robust operation across diverse conditions. The robot employs soft, twisted beams with complex dynamics previously explored in~\cite{twist_beam}.

Leveraging the frequency-dependent dynamic response of its legs, Flix-Walker achieves a number of effective locomotion degrees of freedom that exceeds its actuator count by employing dynamics-based demultiplexing. In this framework, the planar vibrational force input — characterized by shaking frequency ($f$) and direction ($\theta$) — serves as a composite control signal that excites the robot’s longitudinal, lateral, and rotational motion components (see~\figureref{fig:concept}(b)).  The vibration frequency and direction act as selectors, while the robot’s intrinsic dynamics function as a demultiplexer, selectively expressing and amplifying specific motion components in response to the chosen $f$ and $\theta$. By fine-tuning these parameters, Flix-Walker can express and amplify various combinations of three distinct locomotion modes: linear translation along its longitudinal axis, turning via coordinated activation of longitudinal and rotational dynamics, and lateral strafing by isolating the lateral motion component. This signal-processing-inspired demultiplexing approach provides a novel strategy for programmable gait generation, enabling tunable transitions between locomotion modes and offering a powerful new paradigm for underactuated robotic mobility.

To analyze the Flix-Walker’s dynamic behavior under various control inputs, we developed a dynamic model based on the pseudo-rigid-body method, which was implemented in the MuJoCo simulator~\cite{mujoco} and calibrated to match the prototype's performance. Through simulations, validated via experiments with the prototype, we have identified effective actuation parameters and examined the system’s sensitivity to system and environmental variation. These findings establish a robust framework for understanding the complex relationships guiding robot behavior and for selecting actuation parameters that are resilient to key sources of variation.

\subsection{Background}
\label{sec:background}

Under-actuated compliant systems are increasingly prominent in soft robotics for reducing actuation and control requirements. Advances in metamaterials and mechanical structures have enabled soft robots to perform sensing~\cite{xia_responsive_2022,jiao_mechanical_2023} and actuation~\cite{bonfanti_automatic_2020,ZHENG2022101662} tasks more efficiently, addressing key challenges in actuation and control. However, conventional soft robots, despite their adaptability and numerous degrees of freedom, usually require bulky actuators and complex controllers to achieve specific locomotion tasks~\cite{li_compliant_2021,chopra_toward_2023}, complicating modeling, fabrication, and scalability.

Legged locomotion has advanced across a broad size range, from meter-scale~\cite{anymal,birdbot} to centimeter- and millimeter-scale robots~\cite{Andrew2014,nyakatura_reverse-engineering_2019,liu_wireless_2024}. While these robots can be highly maneuverable, most use more than ten actuators, increasing power, system complexity, and fabrication costs.

Underactuated locomotion strategies have been developed to minimize actuator count. Some designs use mechanical linkages or body curvature for predefined gaits, enabling high-speed movement with fewer actuators~\cite{faal_hierarchical_2016, feshbach_curvequad_2023}. Recent studies show that leveraging material properties and structural dynamics allows for effective locomotion with fewer actuators~\cite{bhounsule_low-bandwidth_2014, islam_scalable_2022}, enabling simpler, scalable robots for swarm and exploration applications. However, these approaches often restrict robots to limited movement modes—typically linear translation or turning—reducing versatility.

\begin{table}[t]
\centering
\caption{Comparison with State-of-the-Art Vibration-driven Locomotors}
\begin{tabularx}{\columnwidth}{>{\centering\arraybackslash}m{.65cm} X >{\centering\arraybackslash}m{2.4cm} >{\centering\arraybackslash}m{.7cm} >{\centering\arraybackslash}m{.7cm} >{\centering\arraybackslash}m{.85cm}}
\hline
Work Index & Actuator Count & Capable Locomotions & Body Length (mm) & Body Mass (g) & Power Source \\ \hline
This Work  & 2           & Forward locomotion, Turning, Strafing   & 220    &  350   &  Onboard    \\
\cite{Birkmeyer2009} & 1     & Forward locomotion             & 100    &   16.2    & Onboard   \\ 
\cite{zhu_5-mm_2022} & 1          & Forward locomotion, Turning     & 20         &  0.0756     &  Onboard     \\
\cite{tang_multidirectional_2024} & 1  & Forward locomotion, Turning        &  25    &  1.2    &   Onboard    \\
\cite{kilobot_2012} & 2             & Forward locomotion, Turning        & 33         & 18     &  Onboard    \\
\cite{li_agile_2019} & 2        & Forward locomotion, Turning      & 40         &  5     &   Tethered   \\ 
\cite{yan_terrain_2022} & 2     & Forward locomotion, Turning             & 35    &   6    & Onboard   \\  \hline

\end{tabularx}
\label{tab:lit_review}
\end{table}

Vibrational actuation provides an alternative paradigm for achieving complex locomotion with minimal actuators. By centrally applying oscillatory forces, vibrational actuators can distribute energy throughout a robot’s body, simultaneously driving multiple appendages. This approach has found success in millimeter-scale legged robots~\cite{Birkmeyer2009,tang_multidirectional_2024,kilobot_2012,zhu_5-mm_2022,li_agile_2019,yan_terrain_2022}, which utilize stick-slip mechanisms for agile movement. However, vibrational actuation remains relatively unexplored at the centimeter scale, largely due to the energetic constraints of available actuators at this size.

In this context, we introduce Flix-Walker, a centimeter-scale (20\,cm, 350\,g) locomotor that combines dynamically-coupled compliant mechanisms with vibrational actuation. Driven by just two brushless motors and equipped with 3D-printed soft, twisted beams, Flix-Walker achieves three distinct mobility degrees of freedom: forward locomotion, turning, and strafing. This is enabled by dynamics-based demultiplexing, where different vibrational frequencies and directions (serving as selector signals) excite specific gait dynamics within the robot’s compliant structure (acting as a demultiplexer). As a result, Flix-Walker attains more mobility DoF than the number of actuators, while also simplifying manufacturing and reducing costs. To contextualize Flix-Walker’s performance, we compare its locomotion capabilities and body scale to those of other state-of-the-art vibration-driven locomotors (see~\tableref{tab:lit_review}).

This study fits under the umbrella of a new class of devices termed Soft Curved Reconfigurable Anisotropic Mechanisms (SCRAMs), previously explored in pinched tubes~\cite{9341109,9479201,jiang_shape_2021}, buckling beams~\cite{9244584,9369911}, and twisted beams~\cite{twist_beam}. Leveraging the shape and material properties of soft structures allows complex actuation signals to be consolidated and simplified for generating complex motions.

\subsection{Contributions}
\label{sec:contribution}
The contributions of this paper are summarized as follows: 
\begin{enumerate}
    \item This paper introduces a novel underactuated robotic locomotor that uses just two motors to generate vibrational forces, with soft twisted beam limbs enabling tunable dynamic locomotion.
    \item This paper presents a revised Pseudo-Rigid Body Model (PRBM) with an optimization-based calibration process, permitting high-fidelity dynamic simulation of the compliant system in a rigid body simulator (MuJoCo). 
    \item This paper demonstrates how selecting different vibration inputs enables the robot’s intrinsic dynamics to act as a demultiplexer, producing distinct locomotion modes.
    \item This paper investigates the impact of manufacturing variations and environmental disturbances on robot behavior through simulation and experiments. A generalized method is developed to select control parameters for underactuated compliant robots, balancing performance and robustness across various locomotion modes.
\end{enumerate}
\section{Robot Design and Modeling}
\label{sec:design_modeling}
This section presents the design and modeling of Flix-Walker, covering full-robot simulation, pseudo-rigid-body modeling of twisted beam legs, and the actuation system. It describes parameter identification for leg model calibration and uses simulation to analyze actuation dynamics and validate the design.

\subsection{Robot Design and Prototyping}
\label{sec:design}

This section presents the design and fabrication of Flix-Walker with three primary components: 1) a lightweight, rigid carbon fiber body plate for efficient transmission of actuation forces; 2) four soft, twisted-beam legs at the plate’s corners, each terminating in a rigid, 3D-printed PLA foot; and 3) a duo-rotor shaker system at the body center for actuation, as described in \sectionref{sec:actuator} and shown in \figureref{fig:simulation}(d).
\begin{enumerate}
    \item Robot Body: The body is constructed from a 16\,g CNC-machined. carbon fiber sandwich sheet (DragonPlate EconomyPlate Carbon Fiber Balsa Core, 3.5\,mm thick). This design provides the necessary rigidity to transmit vibrational forces while minimizing mass and inertia.
    \item Robot Legs: The legs are 3D printed in TPU95A, following the geometry from~\cite{twist_beam} with a 90$^\circ$ twist angle~$\phi$ and $T_{\text{leg}} = 3$\,mm thickness for effective locomotion. Key parameters are shown in \figureref{fig:simulation}(c) and \tableref{tab:design}.
    \item Robot Prototype: The prototype (see~\figureref{fig:concept}(a)) integrates an ESP32 microcontroller with SimpleFOC motor driver at the tail and a 300\,mAh 3-cell Li-Po battery at the front for balance. The total weight is 350\,g. Two brushless motors with offset rotary loads are mounted in the cage. Full specifications are listed in~\tableref{tab:design}.
\end{enumerate}

\begin{table}[t]
\begin{center}
\caption{Design Parameters}
\setlength{\extrarowheight}{1pt}
    \begin{tabular}{  c  c  c  c  p{5cm} }
    \hline
    Parameter & Symbol & Value & Unit\\
    \hline
    Body Plate Length & $L\_body$ & 220 & mm\\
    Body Plate Width & $W\_body$ & 120 & mm\\
    Body Plate Thickness & $T\_body$ & 3.5 & mm\\
    Leg Length & $L\_leg$ & 50 & mm\\
    Leg Width & $W\_leg$ & 20 & mm\\
    Leg Thickness & $T\_leg$ & 3 & mm\\
    Leg Twist Angle & $\phi$ & 90 & degree\\
    Foot Length & $L\_foot$ & 85 & mm\\
    Foot Width & $W\_foot$ & 5 & mm\\
    Rotor arm length & $l$ & 28 & mm\\
    Rotor vertical distance & $h$ & 28 & mm\\
    Rotor Offset Mass & $m$ & 12 & g\\
    Rotor frequency& $f$ & - & Hz\\
    Rotor offset angle& $\theta$ & - & degree\\
    \hline
    \end{tabular}
    \label{tab:design}
\end{center}
\end{table}

\subsection{Robot Modeling}
\label{sec:simulation_config}

\begin{figure*}[t]
    \centering
    \includegraphics[width = \textwidth]{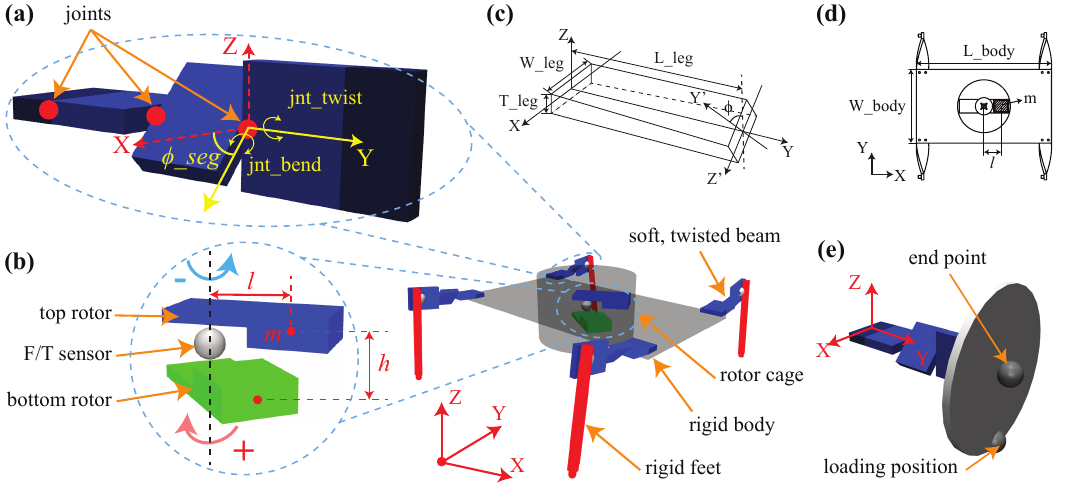}
    \caption{\textbf{Modeling and simulation setup of Flix-Walker.} (a) The simulation model for the duo-rotor actuation system; (b) The pseudo-rigid-body model of the soft twisted beam. (c) Design diagram of the soft twisted beam leg; (d) Design diagram of the robot body. (e) Simulation setup for calibrating the single-leg model. 
    }
    \label{fig:simulation}
\end{figure*}

The dynamic model of the Flix-Walker was developed for simulation in MuJoCo. This comprehensive model integrates a pseudo-rigid-body approach to represent the soft twisted beam design, alongside rigid components that match the physical prototype, as detailed in~\sectionref{sec:design}. An overview of the resulting MuJoCo model is shown in~\figureref{fig:simulation}.

Due to the robot’s complexity — with four twisted beams, four contact points, and two rotors — calibrating the model presents significant challenges, particularly given the large search space of the model's parameters. Further complicating calibration, manufacturing and assembly inconsistencies in the physical prototype introduce additional sources of variability. To improve model fidelity and ensure close alignment with real-world behavior, a two-stage calibration process was implemented using reference data collected from prototype experiments.

This process involves: 1) single leg identification, which establishes baseline stiffness and damping values for the soft twisted beam legs; and 2) full model calibration, which further tunes these baseline parameters and introduces variance factors to account for manufacturing and assembly differences. This two-stage approach enhances the realism and accuracy of the model.

\subsection{Single Leg Identification}
\label{sec:leg_model}
Pseudo-rigid body modeling (PRBM)~\cite{howell_evaluation_1996} is employed to characterize the dynamic behavior of the soft twisted beams, owing to its computational efficiency and compatibility with widely used rigid-body dynamics simulators such as MuJoCo. As illustrated in~\figureref{fig:simulation}(b), our implementation of the PRBM divides the beam into three equal-size, equal-mass rigid segments connected in series. The first segment is aligned with the base, while each subsequent segment is twisted by an offset angle of $\phi_\textit{seg} = 45^\circ$ about its local Y-axis relative to its parent segment. Specifically, the connection between the first and second segment has no twist ($\phi = 0$), while the connections to the third and fourth segments introduce $45^\circ$ twists each, resulting in a cumulative twist of $90^\circ$ between the first and last segments along the initial segment's Y-axis. At each connection between segments, two orthogonal revolute joints are placed: one aligned with the local X-axis to model bending, and the other with the local Y-axis to model twisting. All bending joints share a common set of stiffness and damping parameters, while all twisting joints share a common second set of parameters. This arrangement reflects the beam’s constant width, thickness, and material properties (UltiMaker TPU 95A). The joint stiffness is directly proportional to the beam’s flexural rigidity and inversely proportional to the segment length.

A simulation-based optimization approach is used to identify the single-leg model parameter set $S$, specifically the stiffness ($k$) and damping ($b$) coefficients for the pseudo-rigid bending and twisting joints ($k_{\text{bend}}, b_{\text{bend}}, k_{\text{twist}}, b_{\text{twist}}$).

To estimate these parameters, a series of laboratory experiments is conducted on a soft twisted beam matching the dimensions of the robotic leg described in~\sectionref{sec:design}. In each trial, the beam is laterally deflected, and a load is attached to its distal end. The load is then suddenly released, allowing the beam to return to its original position. The motion of the beam’s distal end is recorded as it recovers. This procedure is repeated under three loading conditions, with the load placed at $-60^\circ$, $0^\circ$ (vertical), and $60^\circ$. The resulting reference data from these tests is used for parameter fitting.

The same loading and unloading protocol is replicated in a MuJoCo simulation, as illustrated in~\figureref{fig:simulation}(e), and the simulated end-point motion during recovery is recorded.

For parameter optimization, a Bayesian hyperparameter optimization algorithm~\cite{Cowen-Rivers2022-HEBO} is employed to minimize the objective function, defined as the root mean square error (RMSE) between simulated and experimental position ($P_i$) and rotation ($R_i$) data. The optimization problem is formulated as:
\begin{equation}
    S^* = \arg\min_{S} 
    \mathrm{RMSE}\left(
        [P_i(n), R_i(n)],
        [\hat{P}_i(n), \hat{R}_i(n)]
    \right),
    \label{eqn:minimization}
\end{equation}
where $S = \{k_{\text{bend}}, b_{\text{bend}}, k_{\text{twist}}, b_{\text{twist}}\}$, and $i \in \{x, y, z\}$ denotes the spatial axes; $P_i(n)$ and $R_i(n)$ are the simulated position and rotation at time step $n$ along axis $i$, while $\hat{P}_i(n)$ and $\hat{R}_i(n)$ are the corresponding experimental reference data. The comparison between reference and model-fitted results is shown in~\figureref{fig:model_data}(a). The optimized parameter set $S^* = \{1.062072, 2.21209, 0.0024, 0.00183\}$.

\subsection{Full Model Calibration}
\label{sec:model_cal}
After identifying the single-leg model, we apply the same fitting method to calibrate the full robot model, addressing discrepancies arising from manufacturing and assembly errors. The error parameter set $\mathbf{X}$ is introduced to capture these variations, defined as:
\begin{align}
\mathbf{X} = \big\{ 
    & \delta_{k,\text{bend},i},\ \delta_{b,\text{bend},i},\ \delta_{k,\text{twist},i},\ \delta_{b,\text{twist},i},\notag \\
    & f_i,\ m_{\mathrm{mag}},\ m_{\mathrm{x}},\ m_{\mathrm{y}} \big\}
\label{eqn:objective}
\end{align}
where $i \in \{\mathrm{fl},\mathrm{fr},\mathrm{rl},\mathrm{rr}\}.$
These parameters represent the proportional errors ($\delta$) in stiffness ($k$) and damping ($b$) for each leg, relative to the baseline values identified in~\sectionref{sec:leg_model}, and are specified for the front-left (fl), front-right (fr), rear-right (rr), and rear-left (rl) feet. Additional parameters account for the body mass ($m$), offsets in the body’s $x$ and $y$ positions, and the tangential friction coefficients ($f$) for each foot.

Experiments are conducted using the prototype, as described in~\sectionref{sec:design} and illustrated in~\figureref{fig:concept}(b), to gather reference data. During these tests, the robot is actuated at various driving frequencies ($f\in[-35,\,35]$ Hz) and shaking orientations ($\theta \in [-90,\,90]^\circ$) for $8$\,seconds, while its pose data is recorded. The same actuation commands are then replicated in simulation, and the resulting pose data is collected for calibration purposes.

\begin{figure*}[t]
    \centering
    \includegraphics[width = \textwidth]{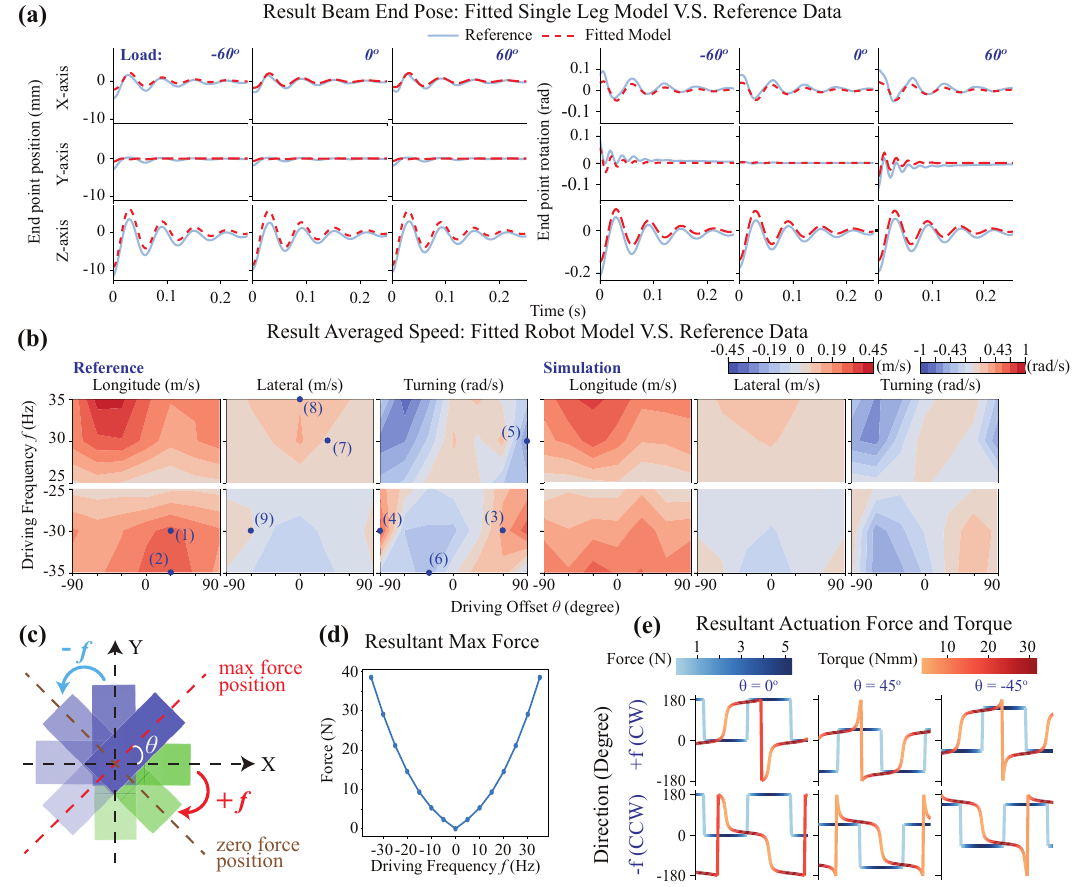}
    \caption{\textbf{Single leg and whole body model calibration result and actuation system simulation result.} (a) Single leg model fitting result comparison using end point position and rotation data. (b) Comparison of whole-body model fitting results with experimental data from the prototype, using averaged linear velocities in the longitudinal and lateral directions and turning velocity. Index numbers indicate the selected actuation pairs for each locomotion mode (see~\sectionref{sec:locomo_mode} for details). (c) Illustration of the actuation system showing its offset angle and driving directions. (d) Simulation result: resultant max actuation force v.s. driving frequency. (e) Resultant actuation force and torque for various driving frequencies ($f = \pm10$\,Hz) and offset angles ($\theta = 0^\circ, 45^\circ, -45^\circ$) within one actuation cycle. The Y-axis shows the direction of the resultant force and torque in the X–Y plane, while the color indicates the corresponding magnitude.
    }
    \label{fig:model_data}
\end{figure*}

The same Bayesian hyperparameter optimization algorithm is used to minimize an objective function that quantifies the discrepancy between the robot’s simulated and experimental motion. Specifically, the objective considers the RMSE between the simulated and measured average linear velocities ($V_i$) in the longitudinal ($x$) and lateral ($y$) directions, as well as the angular velocity ($\omega$) about the Z-axis:
\begin{equation}
    \mathbf{X}^* = \arg\min_{\mathbf{X}} 
    \mathrm{RMSE}\left(
        [V_i(n), \omega(n)],
        [\hat{V}_i(n), \hat{\omega}(n)]
    \right),
    \label{eqn:minimization_robot}
\end{equation}
where $i \in \{x, y\}$ denotes the spatial axes, $V_i(n)$ is the simulated linear velocity along axis $i$ at time step $n$, $\hat{V}_i(n)$ is the corresponding experimental value, $\omega(n)$ is the simulated angular velocity about the $z$-axis, and $\hat{\omega}(n)$ is the experimental reference. The average speed from both the prototype experiments and the simulation, using the optimized parameters, is presented in~\figureref{fig:model_data}(b). The simulated motion closely matches the reference data, indicating the model accurately captures the robot's dynamics.
\subsection{Actuation Analysis}
\label{sec:actuator}
Flix-Walker employs a dual-motor system to generate a unidirectional vibrational force, as illustrated in~\figureref{fig:simulation}(a). Two motors, each with an off-axis mass $m$ mounted at an arm length $r$, are positioned at the center of the body and rotate in opposite directions about a shared axis. The vertical separation between the two offset masses is $h$. This arrangement produces a consistent shaking force in a single direction while largely canceling out rotational inertia.

As shown in~\figureref{fig:model_data}(c), the centrifugal forces produced by both masses add constructively along the axis of shared motion and cancel along the orthogonal axis. The angle $\theta \in [-90, 90]$,degrees defines the offset between the robot’s sagittal plane and the direction of maximum force, with positive $\theta$ to the right and negative to the left. When $\theta = 0^\circ$, the force is directed along the forward ($X$) axis; when $\theta = \pm90^\circ$, it is aligned with the lateral ($Y$) axis.

The driving frequency $f$ sets the rotational speed of the top rotor: positive values correspond to clockwise rotation, negative to counterclockwise. The bottom rotor spins at the same speed but in the opposite direction, effectively minimizing net rotary inertia. The position vectors of the two rotors are:
\begin{equation}
\vec{r}_1(t) = 
\begin{bmatrix}
l \cos(\omega t + \theta) \\
l \sin(\omega t + \theta) \\
h/2
\end{bmatrix},\
\vec{r}_2(t) = 
\begin{bmatrix}
l \cos(-\omega t + \theta) \\
l \sin(-\omega t + \theta) \\
-h/2
\end{bmatrix},
\end{equation}
where $\omega = 2\pi f$.
The centrifugal force generated by each rotor is:
\begin{equation}
\vec{F}_1(t) = m\omega^2 l 
\begin{bmatrix}
\cos(\omega t + \theta) \\
\sin(\omega t + \theta) \\
0
\end{bmatrix},\
\vec{F}_2(t) = m\omega^2 l 
\begin{bmatrix}
\cos(-\omega t + \theta) \\
\sin(-\omega t + \theta) \\
0
\end{bmatrix}.
\end{equation}
The resulting net force, including gravity, is:
\begin{equation}
\begin{aligned}
\vec{F}_{\text{total}}(t) &= \vec{F}_1(t) + \vec{F}_2(t) + \vec{F}_g \\
&= 2m\omega^2 l \cos(\omega t)
\begin{bmatrix}
\cos\theta \\
\sin\theta \\
0
\end{bmatrix}
+
2m\begin{bmatrix}
0 \\
0 \\
-g
\end{bmatrix}.
\end{aligned}
\label{eq:force}
\end{equation}
The net torque generated, including gravity, is:
\begin{equation}
\begin{aligned}
\vec{\tau}_{\text{total}}(t) &= \vec{r}_1(t) \times \big[\vec{F}_1(t) - mg\hat{\mathbf{z}}\big]
+
\vec{r}_2(t) \times \big[\vec{F}_2(t) - mg\hat{\mathbf{z}}\big] \\
&= \begin{bmatrix}
- h m \omega^2 l \sin(\omega t) \cos\theta
- 2 m g l \cos(\omega t) \sin\theta \\
- h m \omega^2 l \sin(\omega t) \sin\theta
+ 2 m g l \cos(\omega t) \cos\theta \\
0
\end{bmatrix}.
\end{aligned}
\label{eq:torque}
\end{equation}

As shown in~\equationref{eq:force}, the dual-motor system generates the intended unidirectional shaking forces. Although the $Z$-axis moment is canceled (\equationref{eq:torque}), a residual shaking torque remains in the $X$-$Y$ plane, varying with frequency ($f$) and direction ($\theta$) due to rotor momentum exchange. The vertical rotor separation $h$ amplifies this torque, reducing symmetry about the Sagittal and Coronal planes. This dynamic torque causes the asymmetric locomotion seen in the speed profiles of~\figureref{fig:model_data}(b).

The actuation system is further analyzed in MuJoCo to study the dynamic forces and torques it generates. The simulation setup, shown in~\figureref{fig:simulation}(a), closely replicates the prototype using design parameters from \tableref{tab:design}. Two rotors are attached to a rotational joint equipped with force and torque sensing. During simulation, the rotors are driven at specified frequencies and offset angles $\theta$, allowing the resulting forces and torques to be recorded.

The directions of the resultant force and torque for driving frequencies $f = \{-10, 10\}$\,Hz and offset angles $\theta = \{-45, 0, 45\}^\circ$ are presented in~\figureref{fig:model_data}(e), demonstrating how control inputs produce targeted shaking forces. The results confirm that the actuation system can generate unidirectional shaking forces as intended, with the offset angle $\theta$ effectively setting the force direction. \figureref{fig:model_data}(d) shows how the maximum shaking force scales with driving frequency. Detailed simulation results are provided in the supplementary video\,1. 

\section{Mobility and Sensitivity Analysis}
\label{sec:locomotion}
Results from both the prototype and simulation, as shown in~\figureref{fig:model_data}(b), indicate that actuation settings play a critical role in determining the locomotion modes of Flix-Walker. They also illustrate the chaotic transitions in resulting speeds with respect to actuation conditions, highlighting the high sensitivity of Flix-Walker’s locomotion to variations in both the system and its environment.

This section first analyzes the experimental data from the prototype, classifying the various locomotion modes of Flix-Walker and illustrating how actuation settings can alter the resulting locomotion mode. Subsequently, we leverage the established simulation as a powerful tool to comprehensively investigate the system’s sensitivities to both internal and external variation factors.

\subsection{Mobility Analysis}
\label{sec:locomo_mode}
\begin{figure*}[t]
    \centering
    \includegraphics[width = \textwidth]{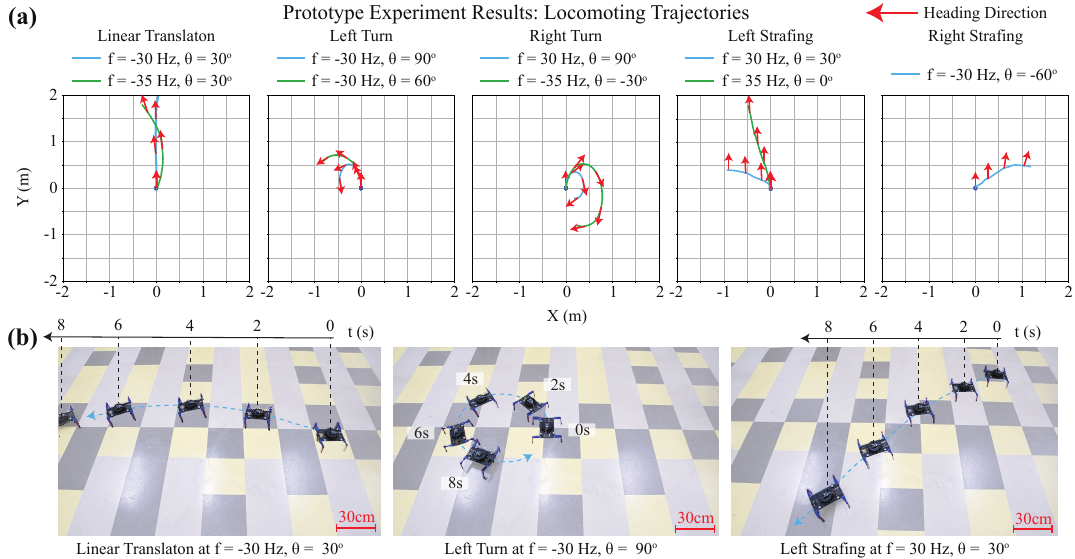}
    \caption{\textbf{Locomotion Analysis} (a) Selected resulting trajectories from prototype experiments demonstrating various locomotion modes; (b) Time-lapse of example locomotion modes (left to right: linear translation, left turn, and left strafing) obtained from the lab experiments.
    }
    \label{fig:locomo}
\end{figure*}

We explored how Flix-Walker’s different locomotion modes emerge by conducting a series of prototype experiments under various actuation settings. As described in~\sectionref{sec:model_cal} and illustrated in~\figureref{fig:model_data}(b), we systematically mapped each combination of vibration frequency ($f$) and offset angle ($\theta$) to its resulting locomotion behavior. The experimental trajectories were examined and grouped according to their motion characteristics, as shown in~\figureref{fig:locomo}(a), summarized in~\tableref{tab:locomo}, and highlighted with selected examples in~\figureref{fig:locomo}(b).

As expected for a robot moving in a 3D space under a planar motion constraint, Flix-Walker possesses three degrees of freedom: two for translation in the plane and one for rotation. When actuated by the shaker, the robot’s response blends these degrees of freedom, with the proportions determined by the frequency $f$ and direction $\theta$ of the vibration input. Each locomotion mode exhibits a characteristic velocity profile: linear translation is dominated by forward motion with minimal lateral or rotational movement (e.g., points (1,2) in~\figureref{fig:model_data}(b)); turning is characterized by pronounced rotation with low forward and lateral velocities (points (3,4,5,6)); and strafing emphasizes lateral movement, with little forward or rotational speed, allowing the robot to move sideways while maintaining a fixed heading (points (7,8,9)). These findings demonstrate that Flix-Walker’s underactuated design can reliably produce a wide variety of useful and effective movement patterns with few actuators.

\begin{table}[t]
\centering
\caption{Locomotion Mode Table}
\begin{tabular}{>{\centering\arraybackslash}m{2.5cm} >{\centering\arraybackslash}m{1.5cm} >{\centering\arraybackslash}m{1.5cm} >{\centering\arraybackslash}m{1.5cm}}
\hline
Locomotion Mode & Actuation Frequency ($f$\,Hz) & Actuation Offset ($\theta^o$) & Indices in \figureref{fig:model_data}(b)\\ \hline
Linear Translation I  & -30         & 30  & (1) \\ 
Linear Translation II  & -35         & 30  & (2) \\ 
Left Turn I         & -30        & $\pm$90  & (3) \\
Left Turn II         & -30        & 60  & (4) \\
Right Turn I       & 30         & $\pm$90  & (5) \\ 
Right Turn II       & -35         & -30  & (6) \\ 
Left Strafing I  & 30         & 30   & (7)  \\
Left Strafing II & 35         & 0   & (8)  \\
Right Strafing & -30        & 60  & (9) \\ \hline
\end{tabular}%
\label{tab:locomo}
\end{table}

\subsection{Sensitivity Analysis}
\label{sec:dynamic_simulation}
\begin{figure*}[t]
    \centering
    \includegraphics[width = \textwidth]{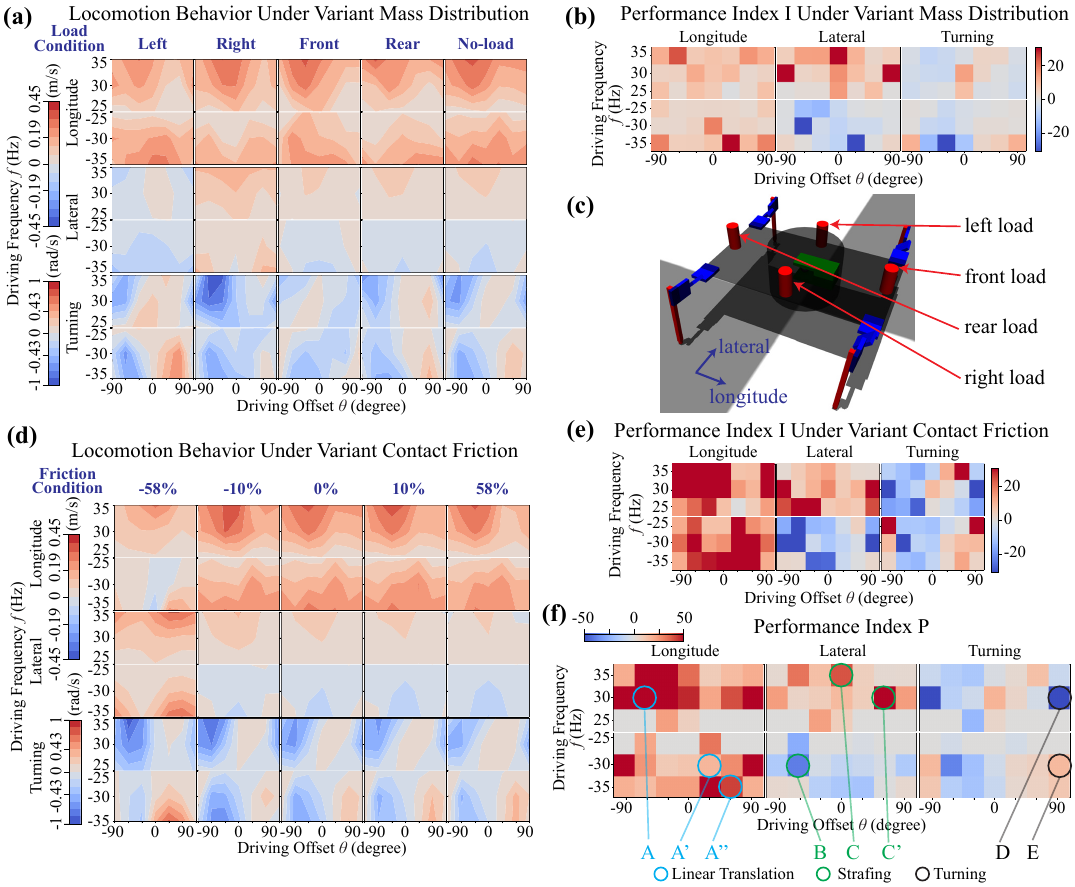}
    \caption{\textbf{Sensitivity analysis using simulation.} (a) Simulation results from various mass distribution conditions; (b) actuation performance index $I$ under mass distribution variants; (c) MuJoCo model for simulation of different load conditions; (d) simulation results from various contact frictions; (e) actuation performance index I under contact friction variants; (f) actuation robustness index ($P$) map with selected actuation pairs.
    }
    \label{fig:sensitivity}
\end{figure*}

Prototype experiments, as described in~\sectionref{sec:model_cal}, reveal that Flix-Walker frequently exhibits abrupt transitions in locomotion behavior, highlighting the dynamic instability of this under-actuated system. Such sensitivity to variations in both system and environmental factors presents significant challenges in selecting appropriate actuation parameters.

Our study identifies two major factors that significantly influence Flix-Walker’s locomotion: mass distribution differences arising from manufacturing and assembly variations, and contact friction differences due to environmental conditions. We employ a simulation-based approach using MuJoCo with the model described in~\sectionref{sec:simulation_config} to analyze the ranges of actuation parameters — specifically, driving frequency and shaking orientation — that are most resilient to these variations. This understanding enables us to identify combinations of actuation parameters that consistently and robustly produce dynamic motions with minimal variability, even in the presence of internal differences.

\subsubsection{Mass distribution variations sensitivity analysis}
\label{sec:simulation_manufacture_error}

A series of simulations was conducted under various mass distribution conditions. This involved adding an extra 50\,g load to different locations on the robot body, specifically along the front, rear, left, and right edges of the main body, with their centers 12\,mm from the edge, as illustrated in \figureref{fig:sensitivity}(c). This scale of variation was chosen to represent a payload whose center of mass is poorly aligned with that of the robot, reflecting a scenario that might reasonably be encountered in the field. Actuation parameters were systematically varied across a range of values: $f$ from $-35$ to $35$\,Hz and $\theta$ from $-90^\circ$ to $90^\circ$. Each data set was obtained from an 8-second simulation. The averaged speed data, along with baseline simulation results for reference, are shown in \figureref{fig:sensitivity}(a).

Varying the payload distribution clearly impacts the robot’s behavior across all motion directions. Adding mass to the left or right side leads to notable changes in predicted speeds; for example, comparing the longitudinal speed at actuation pairs $(f=-35\,\text{Hz},\ \theta = 60^\circ)$ and $(f=35\,\text{Hz},\ \theta = -60^\circ)$ shows that left-side loading produces a peak longitudinal speed at $\theta = 60^\circ$, while right-side loading removes this peak, disrupting the baseline relationship between longitudinal speed and $\theta$. For lateral speed, under $(f=-35\,\text{Hz},\ \theta = 0^\circ)$ and $(f=35\,\text{Hz},\ \theta = 0^\circ)$, left-side mass increases lateral speed in the negative lateral-axis direction (causing leftward drift), while right-side mass increases speed in the positive lateral-axis direction (rightward drift). Turning speed is also notably affected by lateral mass distribution: with actuation $(f=-30\,\text{Hz},\ \theta=-60^\circ)$, left-side mass increases turning speed in the negative Z-axis direction, while right-side mass increases turning speed in the positive Z-axis direction.

Adding mass to the front or rear of the robot has minor effects on longitudinal and lateral speeds, but turning speed and longitudinal speed can change more substantially. Comparing turning speed data under actuation parameters $(f=-35\,\text{Hz},\ \theta=-30^\circ)$ and $(f=35\,\text{Hz},\ \theta=30^\circ)$ shows that adding mass to either the front or rear can notably increase turning speed in both directions.

Beyond understanding how mass distribution variations affect motion, we aim to identify actuation parameter pairs that yield robust motion outcomes despite these errors. To achieve this, we analyze both the speed deviation and the effectiveness of each actuation pair under mass variations, utilizing an actuation performance index. For each direction, the actuation performance index $I$ is defined as:
\begin{equation}
\label{eqa:index}
I = \frac{V_{\text{ref}}}{\mathrm{RMSE}(V_{\text{load}}, V_{\text{ref}})},
\end{equation}
where $V_{\text{ref}}$ is the reference speed (without additional mass), and $\mathrm{RMSE}(V_{\text{load}}, V_{\text{ref}})$ is the root mean squared error between the velocities with various mass loading conditions ($V_{\text{load}}$) and the reference. This metric highlights actuation commands ($f$, $\theta$) that achieve both high speed and resilience to mass distribution variations. The results of this analysis are shown in \figureref{fig:sensitivity}(b).

\subsubsection{Contact friction variations sensitivity analysis}
\label{sec:simulation_environment_error}

Besides variation from mass distribution, contact friction also plays a significant role in Flix-Walker’s dynamic behavior. We conduct a series of simulations using the setup shown in \figureref{fig:simulation}. The contact model includes two-directional tangential friction, two-directional rolling friction, and one-directional twisting friction; among these, tangential friction primarily influences the motion. We adjust each leg's tangential friction coefficient using calibrated, non-bias friction factors $(f_{\text{fr}}, f_{\text{fl}}, f_{\text{rr}}, f_{\text{rl}}) = (0.508, 0.646, 0.553, 0.451)$, as determined in \sectionref{sec:model_cal}. 

To investigate the impact of asymmetric friction, we systematically vary the tangential friction coefficients of only the left-side legs ($f_{\text{fl}}$, $f_{\text{rl}}$) by $\pm10\%$ and $\pm58\%$. The $-58\%$ offset specifically reflects a condition where the footpad is removed in the prototype, reducing the friction coefficients of the left legs from $(0.646, 0.451)$ to $(0.271, 0.189)$. The results are illustrated in~\figureref{fig:sensitivity}(d).

For longitudinal speed, lower friction on the left side results in more asymmetrical behaviors, particularly at actuation pairs ${f=35\,\text{Hz},,\theta=-60^\circ}$ and ${f=-35\,\text{Hz},,\theta=60^\circ}$, while higher friction restores symmetry. Lateral speed increases on the side with lower friction, shifting the peak direction: a $-58\%$ left-side friction offset produces a positive Y-axis speed peak at ${f=35\,\text{Hz},,\theta=0^\circ}$, while a $+58\%$ offset produces a negative Y-axis peak at ${f=-35\,\text{Hz},,\theta=0^\circ}$. Turning speed also responds to friction asymmetries. A $-10\%$ friction error on the left increases turning speed about the positive z-axis at ${f=30\,\text{Hz},,\theta=60^\circ}$, while the same error on the right increases turning speed about the negative z-axis at ${f=-30\,\text{Hz},\theta=-60^\circ}$.

Finally, to identify actuation parameter pairs that are robust against side friction errors, we apply the same performance index as in~\equationref{eqa:index} to friction variation, with results shown in \figureref{fig:sensitivity}(e).

\subsection{Metric for Selecting Actuation Pairs}
\label{sec:metric}
To identify actuation parameters that generate the desired movements while remaining resilient to error conditions, an index $P$ is introduced to evaluate the overall effectiveness and robustness of each actuation pair. Actuation pairs that fail to produce adequate forward locomotion—specifically, those with a heading speed below 0.05\,m/s—are first excluded. The actuation performance index $P$ is then calculated as the average of the summed indices $I'$ from both the mass distribution error analysis and the friction error analysis:

\begin{equation}
P = \frac{I'{\text{mass}} + I'{\text{friction}}}{2}.
\end{equation}

This index yields high values only when both speed is high and error is low. \figureref{fig:sensitivity}(f) visualizes the results, where darker colors indicate better performance, and red and blue represent the positive and negative direction of the resultant speed, respectively.

Focusing first on linear translation, high longitudinal speed with minimal lateral and turning speeds is prioritized. Based on the metric, three actuation pairs, labeled $A$, $A'$, and $A''$, are identified as optimal candidates. For strafings, the metric favors high lateral speed with low longitudinal and turning speeds. Accordingly, candidates $B$, $C$, and $C'$ are selected: pair $B$ provides a negative Y-direction lateral speed (rightward strafing), while $C$ and $C'$ deliver a positive Y-direction lateral speed (leftward strafing). For turning locomotion, the preferred actuation pairs exhibit high turning speed, lower longitudinal speed, and near-zero lateral speed. Candidates $D$ and $E$ are chosen: pair $D$ provides a negative Z-direction turning speed (rightward turning) with low lateral and favorable longitudinal speeds, while $E$ offers a positive Z-direction turning speed (leftward turning).

\begin{table}[t]
\centering
\caption{Selected Actuation Pairs}
\begin{tabular}{>{\centering\arraybackslash}m{2.2cm} >{\centering\arraybackslash}m{1.5cm} >{\centering\arraybackslash}m{1.5cm} >{\centering\arraybackslash}m{1.5cm}}
\hline
Locomotion Mode & Actuation Frequency ($f$\,Hz) & Actuation Offset ($\theta^o$) & Indices in \figureref{fig:sensitivity}(f)\\ \hline
Linear Translation  & -30         & 30                & A'  \\ 
Left Turn         & -30        & $\pm$90           & E  \\
Right Turn        & 30         & $\pm$90           & D  \\ 
Left Strafing  & 30         & 30                  & C'  \\
Right Strafing & -30        & -60                & B \\ \hline
\end{tabular}%
\label{tab:control}
\end{table}

\section{Working Demonstrations}
\label{sec:demos}

This section presents two untethered closed-loop control demonstrations to illustrate the mobility of Flix-Walker in lab environments. We devised two untethered maneuver tasks using a closed-loop controller with the selected actuation parameters as shown in \tableref{tab:control}.

\subsection{Closed-loop Trajectory Tracking}
\begin{figure}[t]
    \centering
    \includegraphics[width = \columnwidth]{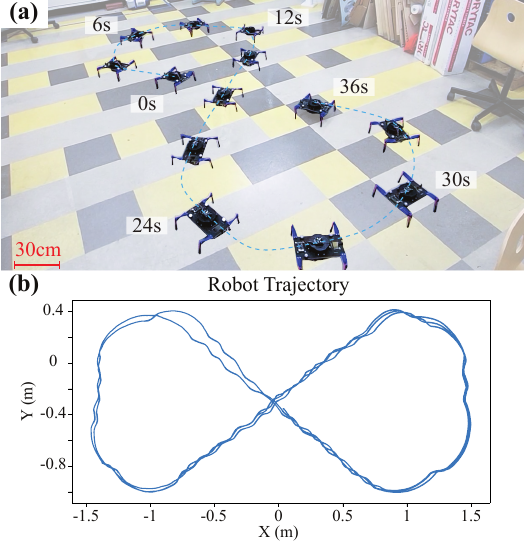}
    \caption{\textbf{Automatic trajectory tracking of a figure '8' pattern.} (a) Time-lapse of the robot trajectory; (b) Robot position data from Optitrack system.
    }
    \label{fig:demo_eight}
\end{figure}

In this demonstration, we implemented a closed-loop, two-state switching feedback controller utilizing turning control parameters derived from \tableref{tab:control} to achieve an '8'-shaped trajectory tracking task. The outcome is shown in \figureref{fig:demo_eight}. During the experiment, the Flix-Walker operates wirelessly while its position is tracked by an Optitrack system. The robot’s pose data is transmitted to a computer, which computes the appropriate control signal and sends it to the robot in real time via WiFi. The controller evaluates the error between the robot’s current heading angle and the desired heading, and applies the corresponding control input based on this error.

The trajectory recorded by the Optitrack system is shown in \figureref{fig:demo_eight}(b). It is evident that the robot closely follows the desired path. A recording of this demonstration is available in the supplementary video\,2.

\subsection{Autonomous Return-to-Origin under Disturbance}
We further demonstrate the self-navigating capability of the Flix-Walker by tasking the robot to autonomously return to a specified origin using the same switching feedback controller. In this demonstration, the robot was intentionally disturbed and pushed away from its original position. Despite these disturbances, the robot effectively navigated back to the origin by relying on the same closed-loop controller. This experiment highlights the robustness of the chosen actuation parameters, as the robot consistently maintains stable locomotion and reliably returns to its starting point, even when perturbed. A video recording of this demonstration is provided in the supplementary video\,2.

\section{Discussion}
\label{sec:discussion}

This paper presents Flix-Walker, a lightweight, under-actuated quadruped that uses soft, twisted beam legs and a controllable one-directional vibration source to achieve diverse locomotion modes. With fewer actuators, the robot selectively activates dynamic behaviors. Experiments systematically examine how actuation parameters affect motion, revealing a broad spectrum of locomotion behaviors through the blending of forward, lateral, and turning motions across the actuation space. While some parameters yield unstable motions, many enable reliable and robust control despite system or environmental variations. Analysis of mass distribution and friction effects provides valuable insights for selecting actuation parameters that are robust to selected factors.

\subsection{Impact}

To identify suitable actuation parameters, we develop a general method for studying SCRAM robots' dynamic behavior. This three-step approach efficiently investigates and selects effective actuation pairs for diverse dynamic motions, supported by a metric designed to identify robust and practical options. The process proceeds as follows:

First, we formulate a simplified model to capture the system's compliant dynamic behavior, calibrating its parameters through prototype testing. Second, we conduct a thorough simulation-based exploration of the actuation domain to analyze dynamic behaviors. Third, we use the proposed metric to identify promising actuation parameter candidates, which we then validate and refine through further prototype experiments to ensure robust and useful locomotion.

Our approach combines fast, calibrated simulation with conventional prototyping to accelerate the exploration of complex robotic dynamics while reducing experimental effort and costs. This method is applicable to SCRAM robots and other underactuated systems with challenging dynamics, and it supports the development of learning-based controllers by enabling large-scale data generation. It also provides a practical framework for sensitivity and robustness analysis that other researchers can readily adopt.

\subsection{Limitations}
\label{sec:limitation}

While our approach is rapid and effective for understanding complex compliant systems and selecting useful actuation parameters, several limitations remain. The robot’s design is not optimized — key aspects like beam dimensions, which strongly influence dynamic behavior, were not tuned, so overall performance may be suboptimal. Furthermore, although both simulation and prototype experiments demonstrate that Flix-Walker’s locomotion modes can be selected by changing vibration force conditions, the motions in 3 DoF cannot be completely isolated due to the underactuated system’s blended dynamic characteristics that arise from mapping 2D actuation forces to 3D motions. The robot’s motion is also limited by actuator power and its own weight: the chosen motor, while the best available in our lab, is still suboptimal and restricts actuation frequencies to below 35\,Hz, which constrains the variety of behaviors observed in experiments — though simulations suggest higher frequencies could enable new modes, such as backward translation seen at 100\,Hz in simulation but not attainable in hardware. Lastly, our modeling approach focuses on simple, continuous geometries that are well-suited to the pseudo-rigid-body model (PRBM) and allow for computational efficiency, but more complex shapes may not benefit directly and would require further tuning and customization.

\section{Conclusion and future work}
\label{sec:conclusion}

This study introduces Flix-Walker, a centimeter-scale underactuated quadruped that uses soft, twisted beam legs and a single unidirectional vibration actuator to achieve forward, turning, and lateral locomotion with just two motors. Through dynamics-based demultiplexing, vibration frequency and direction act as selectors, allowing the robot’s compliant dynamics to amplify specific motions and achieve more mobility modes than actuators. A simplified pseudo-rigid-body model enables fast, accurate simulation in MuJoCo, while a systematic methodology explores the actuation parameter space to identify robust locomotion settings. Performance metrics guide parameter selection, and optimized parameters enable robust trajectory tracking under disturbances, demonstrating Flix-Walker’s effective closed-loop control.

Future work will address the current challenges and limitations by collecting additional data to mitigate overfitting and refining the model to further reduce the gap between simulation and physical experiments. Further optimization of the robot’s design and actuation system will aim to enhance the power-to-weight ratio and motion performance. Additionally, we plan to explore advanced control methods, such as learning-based strategies and automated decision-making algorithms, to enable Flix-Walker to learn control policies for even more complex and demanding tasks, while also investigating how a broader range of real-world variations may impact the robot’s performance.

\bibliographystyle{IEEEtran}

\end{document}